# A COMPARISON OF DATA AUGMENTATION METHODS FOR TRAINING DEEP NEURAL NETWORKS ON SYNTHETIC APERTURE SONAR

CJ Moore University of Missouri, Columbia, Missouri, USA
GD Vetaw Naval Surface Warfare Center Panama City Division, Panama City, Florida, USA
JM Malof University of Missouri, Columbia, Missouri, USA

## 1 INTRODUCTION

Most recent research on automatic target recognition (ATR) in synthetic aperture sonar (SAS) has focused on deep neural networks (DNNs), owing to their tendency to achieve greater accuracy than other approaches[1,2,3,4,5], and in some cases even human experts[7]. One challenge with training DNNs for SAS-ATR, however, is their need for large quantities of labeled training data. While SAS snippets of seafloor are abundant, snippets of target objects are typically scarce due to the high cost of collecting data. One widely used strategy to address data scarcity is augmentation, wherein the goal is to create novel data representations by applying transformations to existing data. The challenge of augmentation is to identify transformations that introduce significant variability while ensuring that the augmented data is realistic. If the augmentations are unrealistic, or add insufficient variability, they can be unhelpful, or even detrimental, when used for training.

The computer vision community has developed a variety of effective augmentations for natural imagery (e.g., imagery from ground-based optical cameras), which we term CV-augmentations, such as rotation, resizing, color jittering, and more[8]. Due to the phenomenological differences between natural imagery and SAS data, it is unclear whether CV-augmentations will work well for SAS[6,7]. This has motivated the investigation of some so-called *physics-informed* (PI)-augmentations that preserve the geometry of a SAS scene. Examples include flipping an image along the platform track[6,10], and photometric transformations[10].

Prior work on the augmentation of SAS data investigated several augmentations[5,10,11]. The most recent work on this topic evaluated the impact of including augmentations one at a time[11]. They found several PI augmentations (e.g., along-track flipping, small rotations, and contrast shifting) and CV augmentations (e.g., zoom) improved DNN recognition accuracy. This study provides valuable evidence that some PI and CV augmentations are beneficial, however, it also suffers from some important limitations. It remains difficult to ascertain whether the authors performed hyperparameter optimization prior to evaluating their augmentations. Most augmentations include hyperparameters that can significantly affect their behavior. Once optimized, the augmentations in previous studies may offer greater benefits overall, and relative to each other. The authors also utilize a relatively small DNN without any pre-training. It is yet to be determined whether these augmentations still provide value for training modern networks, such as transformers, especially those with larger capacity that have been pretrained on optical imagery (e.g., ImageNet[20]).

**Contributions of this work.** In this work we perform a large comparison of both CV and PI augmentations. Compared to prior research, we use a broader set of CV augmentations, including several recent augmentations specifically designed for DNNs, such as Mixup[12], ResizeMix[13], and CutMix[14]. And we also optimize the hyperparameters for each augmentation to better estimate its full potential benefits, if any. In contrast to prior work, we also investigate whether the benefits of the augmentations are additive: to what extent can we combine augmentations to achieve greater benefits than any one augmentation alone? We use a greedy sequential forward search algorithm to sequentially select and add augmentations during training. All of these experiments are conducted using two large widely used DNN-based models: one convolutional model (ResNet[16]), and one transformer-based model (SWIN[15]). We investigate how well the augmentations generalize across different modern DNN architectures; specifically, we optimize our augmentations on one DNN architecture (the SWIN) and then examine whether they are still beneficial when utilized with another architecture (the ResNet). There has been little investigation of transformer-based models for SAS-ATR, and our study provides preliminary insights into their efficacy for ATR, as well as the impact of augmentations for large modern models, such as SWIN.

Our contributions are as follows:
- An analysis of PI and CV augmentations with optimized hyperparameters for SAS-ATR.
- An investigation into effective augmentation policies for SAS-ATR.
- A preliminary comparison of network architectures and their interactions with different augmentations of SAS data as well as of their performance in SAS-ATR more broadly.

# 2 METHODS

## 2.1 Augmentations

In this work, we investigate the effect of twelve different augmentations on DNN performance. We included all augmentations from prior work on SAS data augmentation[11], except those clearly found to be unhelpful. We also included several promising recent augmentations from CV literature: Mixup[12], CutMix[14], and ResizeMix[13]. Below we briefly describe each augmentation. **Table 1** lists all the augmentations as well as the names of the hyperparameters that we vary for each. We use the hyperparameter names as they appear in the function signature of the software libraries that we used (e.g., Torchvision). In this table we also identify which augmentations are considered PI as well as where they appear in the literature. Each augmentation we use can be found in Torchvision[18] except ResizeMix which is available in MMPretrain[17].

| Augmentation | Key Hyperparameter(s) | Appearance in SAS/SAR Literature |
|---|---|---|
| **Along-Track Flip*** | N/A | Orescanin et. al.[11]<br>Williams[6,7] |
| **Contrast Shift*** | **contrast_factor** - contrast scaling factor | Orescanin et. al.[11] |
| **Color Jitter*** | **contrast** – range of scaling factors for random contrast shifts<br>**brightness** – range of scaling factors for random brightness shifts | N/A |
| **Speckle Noising*** | **rate** – controls decay rate of exponential distribution from which noise is sampled<br>**filterSize** – defines size of square median filter | Ding et. al.[9] |
| **Gaussian Blur*** | **kernel_size** – size of blur kernel<br>**sigma** – standard deviation of kernel | N/A |
| **Zoom** | **scale** – range of potential scaling | Orescanin et. al.[11] |
| **Center Shift** | **translate** – range of translation in x and y axes | Orescanin et. al.[11]<br>Williams[7]<br>Ding et. al.[9] |
| **Rotation** | **degrees** – range of potential clockwise and counterclockwise rotations. | Orescanin et. al.[11] |
| **Mixup** | **alpha** – controls mixing ratio | N/A |
| **ResizeMix** | **alpha** – controls mixing ratio | N/A |
| **CutMix** | **alpha** – controls mixing ratio | N/A |

**Table 1:** *Each augmentation we use, the parameters iterated over in optimization, its appearance in SAS/SAR literature, and our source for implementation. Physical augmentations are denoted by an asterisk (*).*

**Zoom** and **center shift** are augmentations that change an image's scale and alignment. They are both affine transformations that use hyperparameters to define a range from which the magnitude of each transformation is sampled. For **zoom**, the "scale" hyperparameter takes a range of potential scaling factors (e.g., [0.8, 1.2] corresponding to a zoom of 80-120% of the original image size), and for **center shift** the "translate" hyperparameter is used to define the range of potential shifts in the x/y direction with respect to image size (e.g. [0.3,0.1] would define a horizontal range of up to ±30% of the number of pixels in the x-axis and a vertical range of up to ±10% of the y-axis from which a shift could occur). Both augmentations sample uniformly from the defined ranges and use black padding in cases where images are translated or scaled down.

**Rotate** and **along-track flip** are affine transformations that manipulate the images spatial orientation. **Rotate** uses a "degrees" hyperparameter to define the maximum degrees of rotation (i.e. [-20,20] means the image can rotate up to 20 degrees in either direction). This range is sampled uniformly and zero padding is also used. **Along-track flip** is performed by setting the "direction" hyperparameter of RandomFlip to the same direction of the sonar platform track ("horizontal" if the track is along the x-axis, "vertical" if along the y-axis). This reflects the image over the axis perpendicular to the track, producing a different view of the image while preserving essential geometry.

**Speckle noise** has also been found to have a significant impact on sonar image quality[19]. To model this noise as an augmentation we follow a prescribed technique from Synthetic Aperture Radar (SAR) research[7] in which a median filter is applied to an image followed by the random sampling of noise from a truncated exponential distribution. This constitutes our **speckle noising** augmentation which we include as well as **Gaussian blur** (as a denoiser) to analyze the effect of noise manipulation as augmentation. The implementation for **Gaussian blur** uses the "kernel_size" and "sigma" parameters to define the blur kernel applied to the image, and **speckle noising** uses hyperparameters "filterSize" to define the size of the median filter and "rate" to set the decay rate of the distribution.

Along with the augmentations that have history in SAS literature we also included some that have seen success in general Computer Vision (CV) research. **Contrast shifting** and **color jitter** are both photometric augmentations that manipulate image pixel intensities. In the context of grayscale images, **color jitter** can only augment the contrast and brightness of an image. Beyond this, these two augmentations differ in that **color jitter** performs its contrast and brightness shifts by randomly sampling scalars from a range of values whereas our implementation of **contrast shift** only uses a constant scalar and does not manipulate image brightness. For **contrast shift** this is controlled by the hyperparameter "contrast_factor" and for **color jitter** the hyperparameters "contrast" and "brightness" accept ranges which are uniformly sampled during training to determine the magnitude of change for contrast and brightness respectively.

Other such augmentations are **ResizeMix**[13], **CutMix**[14], and **Mixup**[12]. **ResizeMix** works by shrinking an image $x_1$ and then overlaying it on a second image $x_2$. The scaling rate of $x_1$, $\tau$, is taken from a uniform distribution defined as $U(\alpha,1.0)$ where $\alpha$ is a user-defined parameter. This rate $\tau$ also defines a "mixing ratio" $\lambda$ (i.e. $\lambda = \tau^2$). This mixing ratio is used to compute a combined image label y' that accounts for the relative sizes of the two combined images ($y' = \lambda y_1 + (1- \lambda) y_2$). This process is similar to that of **CutMix** with the primary differences being how $\lambda$ is computed and how it functions. In **CutMix** the "combination ratio" $\lambda$ is sampled from a Beta distribution ($Beta(\alpha, \alpha)$) as defined using the $\alpha$ parameter. Instead of overlaying a downsized image with another, **CutMix** utilizes a randomly cropped patch. The size of this patch depends on the combination ratio $\lambda$ which also controls the label mixing, using the same formula as **ResizeMix**. **Mixup** is like the other two methods, in that it combines two images, however the way it does so is even simpler. In **Mixup** $\lambda$ is also sampled from the distribution $Beta(\alpha, \alpha)$ and functions identically in label mixing as in the other augmentations. A similar mixing formula is used to combine images $x_1$ and $x_2$ into the image x' (i.e. $x' = \lambda x_1 + (1- \lambda) x_2$). We are unaware of any existing SAS literature that includes these augmentations and given their success in other CV applications we decided to use them in this work.

### 2.2 Network Architectures

The networks evaluated in this work are SWIN-base[15] and ResNet-152[16], both using pretrained ImageNet1k[20] weights. We chose to use these networks because they are relatively close in terms of model capacity, removing this as a potential confounding factor when comparing the performance of the two architectures. Both models also offer strong performance and have an abundance of publicly accessible pre-trained weights. The choice of having one CNN and one transformer-based network was due to the possibility of different interactions between augmentation and architecture. Accounting for these differences will help ensure that no augmentation is unfairly advantaged due to its unique synergy with a particular kind of architecture.

| Network | Type | # Parameters | Citation |
|---|---|---|---|
| ResNet152 | CNN | ~60 million | He et al., 2016[15] |
| SWIN-base | Transformer | ~88 million | Z. Liu et al., 2021[16] |

**Table 2:** *The networks used in this study, whether they are CNN or transformer-based, their parameter count, and the paper introducing them.*

### 2.3 Performance Metrics

To evaluate our networks, we use two performance metrics: F1-score and Average Precision (AP)[21]. F1-score is widely used in SAS-ML literature[7,11]. AP is not as widely used; however, it offers a richer performance characterization than F1 alone as it reflects performance across different decision thresholds on the model output. AP is computed as the mean of all precision values at each recall threshold, effectively representing the area under the PR-curve. This, in conjunction with F1, gives us a richer characterization of network performance by reporting both how our models perform at a single threshold (in the case of F1) as well as across all thresholds (in the case of AP). For this reason, we compare network performance using F1-score and AP.

# 3 EXPERIMENTS

We design our experiments to address several questions: 1) What is the impact on performance of individual augmentations when training DNNs for SAS-ATR 2) Do combinations of augmentations offer greater benefits than individual augmentations? 3) How well do augmentations generalize across models? and 4) How do transformers compare to convolutional models in terms of their performance in SAS-ATR?

**Dataset and Preprocessing:** Our dataset contains 37271 SAS snippets produced by an energy-based prescreener. Of these snippets 1660 are targets and 35661 are clutter (approximately a 1:24 ratio). In our experiments we use randomized two-fold cross-validation where each fold contains approximately half of the total number of snippets and maintains the same class distribution as the larger dataset. During preprocessing, a small Gaussian blur is applied to snippets for denoising. This is followed by min-max normalization and resizing to 256x256px through bilinear interpolation. While these snippets are initially complex-valued, we only use the magnitude data in these experiments.

**Training:** For training we use the MMPretrain package[15] to define our configurations and pipelines. Our networks are trained on clusters of 4 NVIDIA A100X GPUs with 80GBs of VRAM each. Each network uses pretrained ImageNet weights sourced from MMPretrain. We train our networks for 100 epochs using a minibatch size of 64. Our learning rate schedule starts with a linear “warm-up” moving from 3e-5 to 3e-4 over the first 10 epochs followed by cosine annealing for the remainder of training. We use the Adam-W optimizer[22], however for our transformer networks we set the parameter-wise weight decay for positional embeddings to 0 to preserve these embeddings throughout training. Both networks use a focal loss function[23] to help mitigate the drawbacks of training on an unbalanced dataset.

**Augmentation Optimization:** Before we evaluate the extent to which the augmentations we use are beneficial to performance, we set out to determine the best hyperparameters to use for each augmentation. Testing each potential hyperparameter value for every augmentation is intractable, so for each hyperparameter we define 3-4 potential values and use the value that elicits the highest performance. To ensure that each augmentation is given an even comparison in later experiments we chose to only perform this search on a single network, SWIN-b, and use the “best” hyperparameter values found in this search. **Table 3** shows which hyperparameter values we chose for each augmentation this applies to.

| Augmentation | Hyperparameter(s) | Values |
|---|---|---|
| **Zoom** | scale min | **0.5**, 0.65, 0.8 |
| | scale max | 1.2, **1.35**, 1.5 |
| **Center Shift** | translate | 0.1, 0.15, **0.2**, 0.25 |
| **Rotation (small and large)** | magnitude_range | [-10,10], **[-20,20]**, [-30,30], **[-40,40]** |
| **Contrast Shift** | magnitude | 0.25, 0.5, **0.75**, 1.0 |
| **Color Jitter** | brightness | 0.25, 0.5, **0.75** |
| | contrast | 0.25, 0.5, **0.75** |
| **Speckle Noising** | filterSize | **3**, 5, 7 |
| | rate | 2, **3**, 4 |
| **Gaussian Blur** | radius | **3**, 5, 7 |
| **Mixup** | alpha | **0.2**, 0.5, 0.8, 1.0 |
| **ResizeMix** | alpha | **0.2**, 0.5, 0.8, 1.0 |
| **CutMix** | alpha | 0.2, 0.5, **0.8**, 1.0 |

**Table 3:** Augmentations, their hyperparameters and the values we chose for them in bold.

**Impact of Individual Augmentations:** Here we evaluate the impact of individual augmentations. We train the SWIN network once using each of the twelve augmentations, utilizing its optimized hyperparameters from **Table 3.** The results of this experiment are reported in **Figure 1**. We first consider the impact of the augmentations on SWIN (orange bars in **Fig. 1**); recall that the augmentation hyperparameters were optimized for SWIN. The results indicate that, to varying degrees, all augmentations were beneficial for SWIN, utilizing either the F1 or AP performance metric. The Zoom and Shifts were most beneficial, while Mixup was least beneficial. These results suggest that, when their hyperparameters are optimized, these augmentations are unlikely to be harmful and are often beneficial. Next, we investigate whether the augmentations generalize to another model, by inspecting the results obtained with the ResNet (blue bars in **Fig. 1**). The rank-order of the augmentations was similar to the SWIN, but the overall impact of the augmentations is generally larger than before, and some augmentations are detrimental. The best-performing augmentation is again Zoom, and it offers greater performance improvements for the Resnet. However, the Speckle and Mixup augmentations are highly detrimental. These results suggest the augmentations are still usually beneficial, but care must be taken when transferring augmentations from one model to another; our results suggest the results do not always generalize, although it is unclear from our experiments whether this could be remedied by optimizing the hyperparameters on the ResNet.

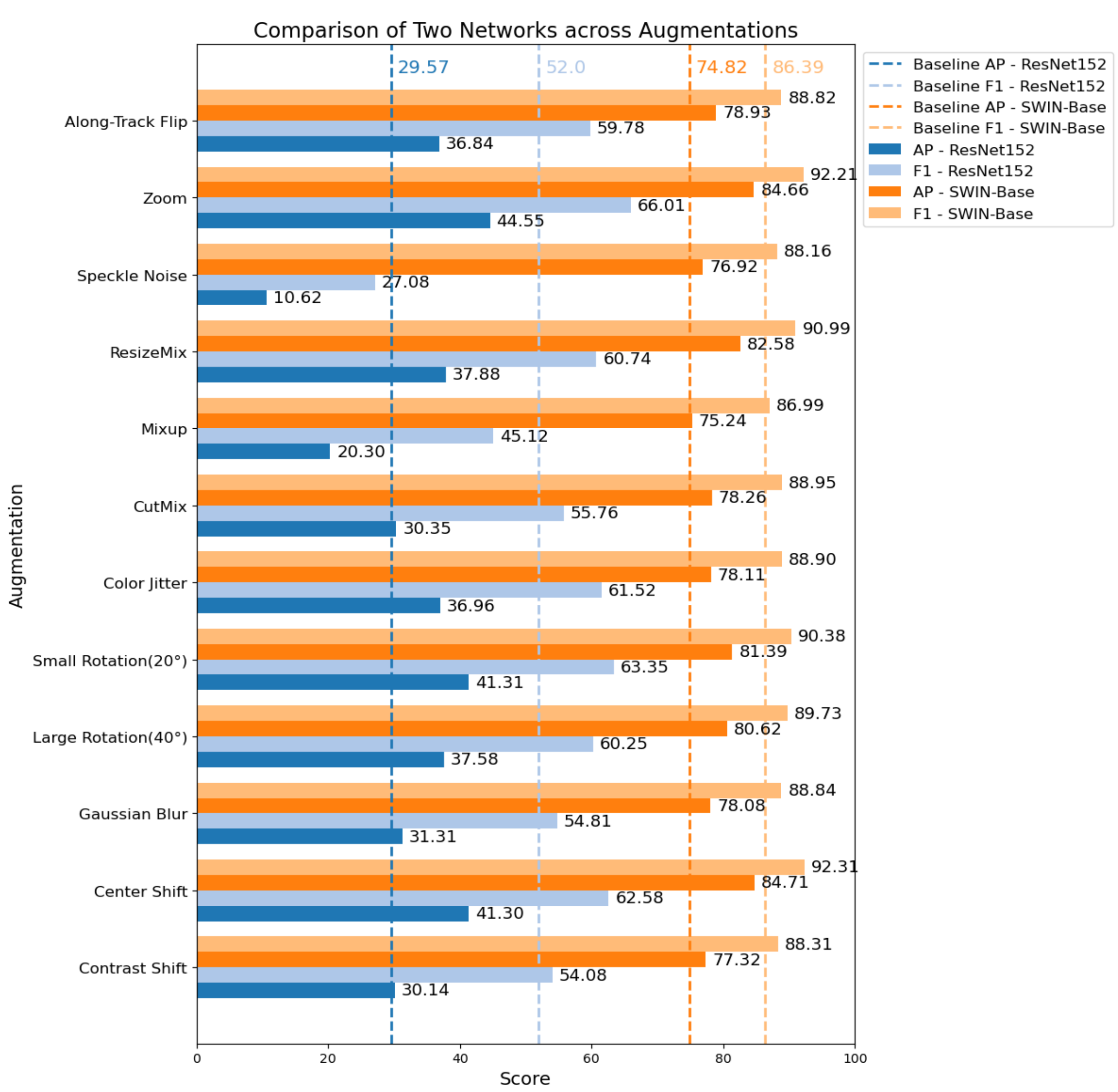

**Figure 1**: AP and F1 scores for each network across all augmentations

## 3.1 Are Augmentations Additive?

To address this question, we perform a greedy selective search to determine a strong multi-augmentation policy for SAS-ATR. This involves constructing an augmentation policy by first testing each augmentation individually and adding the augmentation that elicits the highest performance to that policy. At this point each remaining augmentation is re-evaluated in conjunction with the existing policy and again the one that gives the highest performance is added. This process is repeated until either a) there are no more remaining augmentations or b) performance ceases to improve with new augmentation. We only perform this optimization using the SWIN network, and the results are reported in **Table 4**. Note that the performance of the baseline and zoom varies slightly due to randomness in the training process. The search process completed 4 steps, corresponding to a final policy that comprises four augmentations. The final policy achieves a final AP of 86.33 compared to 84.66 with a single augmentation, and 74.82 for the baseline (no augmentation). These results suggest that multi-augmentation policies are beneficial, but with diminishing benefits as the policy grows. Note that sequential forward search is highly suboptimal, and it is conceivable that better performance could be obtained using a more exhaustive search, but we do not explore that here.

| Greedy Search Step | Augmentation | AP | F1 |
|---|---|---|---|
| Step 0 (baseline) | None | 74.82 | 86.39 |
| Step 1 | Zoom | 84.66 | 92.21 |
| Step 2 | ResizeMix | 86.06 | 92.78 |
| Step 3 | Small Rotation | 86.06 | 92.80 |
| Step 4 | Contrast Shift | 86.33 | 92.92 |

**Table 4**: Our greedy selective search step by step. Which augmentations were chosen at each step and SWIN performance once that augmentation is added to the policy.

As in Sec. 3.1, we take the augmentation policy chosen by SWIN and apply it to the ResNet, to evaluate whether the policy generalizes well. The results are reported in **Table 5** which indicate that it does not offer the same advantages for the ResNet and is similar in performance to the baseline.

| Policy | AP | F1 |
|---|---|---|
| Baseline (no augs) | 29.57 | 52.00 |
| Best Single Augmentation | 44.55 | 66.01 |
| Selective Search Policy | 28.01 | 54.05 |

**Table 5**: ResNet performance with no augmentation, the most helpful single augmentation, and using the policy uncovered in the greedy selective search.

# 4 CONCLUSION

In this work we investigated the potential for DNN performance improvement on SAS-ATR using training data augmentation. We tested both physics-informed (PI) augmentations as well as augmentations from computer vision (CV) research and found that nearly every augmentation we tested was able to produce some increase in AP and F1 for both a SWIN-base and a ResNet152 with the exception of Mixup and speckle noising in the case of ResNet. This suggests that adherence to the physical properties of SAS systems is not required for an augmentation to improve network performance, and in fact, a wide variety of augmentations can do so regardless of their consistency with SAS physics. Additionally, we found that a combination of augmentations (i.e. a policy) can improve performance beyond that of individual augmentations; although an effective policy for one network might not be as effective when applied to another network. This implies that network architecture can have a significant impact on the ability for an augmentation to elicit performance gains. Architecture itself is a major factor in terms of performance as we find that a transformer network (SWIN-base) outperforms a CNN (ResNet152) by substantial margins in all of our experiments. Transformer-based networks have little presence in SAS-ATR literature when compared to their CNN counterparts, however this work demonstrates they show promise in this domain. Future work on this question should involve a more robust analysis of these augmentations across a wide variety of network architectures to determine which architectures benefit most from augmentation in SAS-ATR.

# ACKNOWLEDGEMENTS

This work was supported by the Office of Naval Research (ONR) under grant number N000142412539.